\documentclass[onecolumn]{cohere}

\usepackage{wrapfig}
\usepackage{multirow}
\usepackage{tcolorbox}
\usepackage{enumitem}

\usepackage{pifont}

\definecolor{tabblue}{RGB}{31,119,180}
\definecolor{thomblue1}{RGB}{136,153,244}
\definecolor{thomblue2}{RGB}{83,109,222}

\title{The Multilingual Divide and Its Impact on Global AI Safety}

\multiauthors
\author{name={Aidan Peppin\fa},affiliation={2}}
\author{name={Julia Kreutzer\fa},affiliation={1}}
\author{name={Alice Schoenauer Sebag\fa},affiliation={2}}
\author{name={Kelly Marchisio\fa},affiliation={2}}
\author{name={Beyza Ermis},affiliation={1}}
\author{name={John Dang},affiliation={1}}
\author{name={Samuel Cahyawijaya},affiliation={2}}
\author{name={Shivalika Singh},affiliation={1}}
\author{name={Seraphina Goldfarb-Tarrant},affiliation={2}}
\author{name={Viraat Aryabumi},affiliation={2}}
\author{name={Aakanksha},affiliation={2}}
\author{name={Wei-Yin Ko},affiliation={2}}
\author{name={Ahmet Üstün},affiliation={1}}
\author{name={Matthias Gallé},affiliation={2}}
\author{name={Marzieh Fadaee},affiliation={1}}
\author{name={Sara Hooker\fa},affiliation={1}}
\affiliations{
    \item[1] Cohere Labs
    \item[2] Cohere
}
\corresponding[*]{\{{aidanpeppin}, {juliakreutzer}, {alice}, {kelly}, {sarahooker}\}{@cohere.com}}

\abstract{Despite advances in large language model capabilities in recent years, a large gap remains in their capabilities and safety performance for many languages beyond a relatively small handful of globally dominant languages. This paper provides researchers, policymakers and governance experts with an overview of key challenges to bridging the "language gap" in AI and minimizing safety risks across languages. We provide an analysis of why the language gap in AI exists and grows, and how it creates disparities in global AI safety.  We identify barriers to address these challenges, and recommend how those working in policy and governance can help address safety concerns associated with the language gap by supporting multilingual dataset creation, transparency, and research. 
}

\begin{document}

\section{Introduction}

\begin{quote}
    \textit{The limits of my language means the limits of my world.} \textbf{--- Ludwig Wittgenstein}
\end{quote}

More than 7000 languages are spoken around the world today,\footnote{Eberhard, David M., Gary F. Simons, and Charles D. Fennig. (2024) \href{https://www.ethnologue.com/}{Ethnologue: Languages of the World.} Twenty-seventh edition.}  but current state-of-the-art large language models (LLMs) cover only a relatively small fraction of them. This ``language gap'' has far-reaching implications which ultimately leave certain language communities around the globe marginalized. AI models present both limited language support and biases are introduced that reflect Western-centric viewpoints, undermining other cultural perspectives.

This language gap crucially affects the safety of AI models. Though various efforts have gained traction and momentum around the world to improve the general safety of AI models,  a critical challenge remains: \textit{how to ensure safety across diverse languages and cultures}. This challenge is often widely overlooked or completely absent in efforts to advance AI safety, which primarily focus on English or monolingual settings, leading to potential safety and security flaws for other languages. Part of the problem is the scarcity of reliable datasets for safety evaluation beyond a few languages. Such evaluations are complex and need to reconcile global harms and unique local contexts.

Several research groups have set out to reduce the language and safety gap in AI across diverse linguistic and cultural contexts. One such effort is Cohere Labs’s Aya project\footnote{\url{https://cohere.com/research/aya}} --- a global initiative that has developed and publicly released multilingual language models, instruction datasets, and evaluation datasets expanding language coverage \citep{ustun-etal-2024-aya,dang-etal-2024-rlhf, aakanksha-etal-2024-multilingual,aakanksha_mix_2024,gureja2024mrewardbenchevaluatingrewardmodels,singh2025globalmmluunderstandingaddressing, romanou2025include,dash2025aya}. In the course of this work, many of the challenges and opportunities around expanding the worlds AI serves have become apparent. In this paper, we articulate how these approaches have addressed language disparity and global safety gaps in AI models. This paper is written for both research and policy experts to help provide an overview of the key challenges that remain in bridging the language gap and minimizing safety risks across languages. We provide an analysis of why the language gap exists (Section~\ref{sec:current_state})  and grows (Section~\ref{sec:why}), how it creates gaps in global AI safety (Section~\ref{sec:global}), and an overview of our lab's efforts in context of the Aya intiative (Section~\ref{sec:solutions}), along with technical and fundamental lessons we have learned through this work about how to address the language gap (Section~\ref{sec:recommendations}).

\begin{figure}
    \centering
    \includegraphics[width=0.9\linewidth]{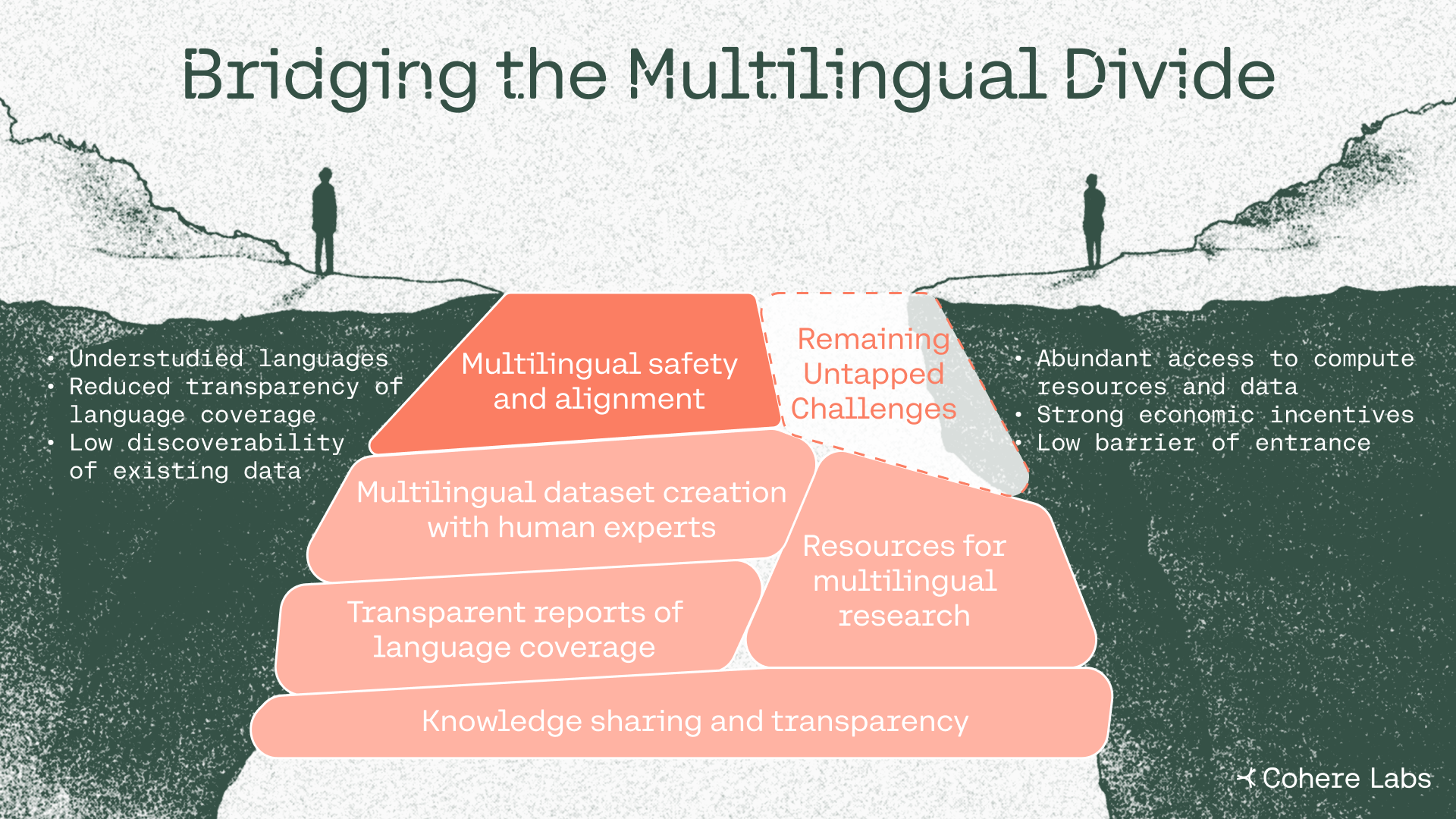}
    \caption{\textit{Bridging the Multilingual Divide: We scrutinize the reasons for the language gap in AI, and review and recommend concrete steps to bridging it. We highlight that the language gap must involve safety mitigation across languages, and that open challenges remain.}
    }
    \label{fig:overview}
\end{figure}

Through this primer, we articulate three overarching barriers that must be overcome to effectively and efficiently close the language disparity and global AI safety gaps for everyone:
\begin{enumerate}
    \item Building high-quality datasets, and curating evaluations using human-curated data from fluent speakers is resource-intensive, but critical for reducing the global AI safety gap.
    \item Access to compute is uneven throughout the work, and is reinforced by disparities in access to digital tools for developing and using large language models.
    \item We must capture not only language but nuances in culture and dialect. Languages are diverse and heterogeneous; while often treated as monoliths, dialects are abundant and regional/cultural nuance must be considered.
\end{enumerate}

In order to overcome these barriers for enhanced global AI safety, we offer the following considerations for policy makers, outlined in the box below.
\begin{tcolorbox}[title=Recommendations for Policy and Research,colback=thomblue1!9]

\begin{enumerate}[left=0pt,label*=\arabic*.]
    \item \textbf{Support multilingual dataset creation:}
    \begin{enumerate}[label*=\arabic*.]
        \item Incentivize and facilitate the creation of open access evaluation sets, which reflect relevant generative use cases and safety-relevant use cases across modalities, by both translating existing datasets ("language-parallel") and creating localized ones ("language-specific").
        \item Fund long-term annotation efforts in endangered languages. This enables human annotators from diverse backgrounds with multilingual and multicultural expertise to engage in the curation of high-quality, inclusive datasets.
    \end{enumerate}
    \item \textbf{Support multilingual transparency from model providers: }
    \begin{enumerate}[label*=\arabic*.]
        \item Encourage model providers to articulate the coverage of languages served by each model family. For example, by reporting languages supported and performance in each language in technical or evaluation reports.
        \item  Conduct analyses of language coverage across safety research, for example by assessing the presence or absence of safety mitigations across languages in published reports.
    \end{enumerate}
    \item \textbf{Support multilingual research and development:}
    \begin{enumerate}[label*=\arabic*.]
        
        \item Support multilingual and non-English research that aims to close the language gap through funding and other programs. 
        \item Enable access to (more) compute for multilingual safety research, especially for projects and in regions where it is disproportionately inaccessible.
    \end{enumerate}
\end{enumerate}
\end{tcolorbox}

\section{State of the Current Language Gap in AI}
\label{sec:current_state}

\begin{tcolorbox}[title=Section Findings,colback=thomblue1!9]
\begin{itemize}[leftmargin=10pt]
\item [\ding{228}] There is a \textbf{significant language gap} in AI development, where the majority of language models are optimized for English and a few other high-resource languages, while many other languages worldwide remain underrepresented. 
\item [\ding{228}] This gap is due to resource disparities, data availability, global inequities, and socio-economic factors, which lead to \textbf{higher costs and limited access} for speakers of low-resource languages. 
\item [\ding{228}] English-centric development and research introduces \textbf{cultural biases }in model outputs, and \textbf{potential safety risks}, exacerbating inequalities and threatening linguistic diversity. 
\item [\ding{228}] Addressing these issues is crucial for ensuring \textbf{equitable access to AI technologies} and preserving cultural representation in the digital age.
\end{itemize}
\end{tcolorbox}

Large language models are finding beneficial applications in a range of contexts across societies and economies around the world. However, the vast majority of language models are currently optimized for a small handful of languages, and the English language and North American socio-cultural preferences are dominant across their design, outputs, and behavior~\citep{yong-etal-2023-prompting, naous_having_2024, cahyawijaya2024univar}. 
 
There are several efforts around the world to develop the multilingual capabilities of AI language models, including Cohere Labs’s Aya models \citep{ustun-etal-2024-aya,aryabumi_aya_2024,dang2024ayaexpansecombiningresearch,dash2025aya} and datasets \citep{singh_aya_2024} --- a family of open source, massively multilingual language models that cover 101 languages, Cohere’s Command A,\citep{cohere2025commandaenterprisereadylarge}, Llama  \citep{dubey2024llama}, Qwen ~\citep{qwen2.5}, Gemma ~\citep{team2024gemma} and Mistral families ~\citep{ministral,mixtral}. Despite concerted research effort, the language gap remains pervasive and models still underperform on languages outside of English \citep{li_quantifying_2024}. 
This language gap in the development, capabilities, and applications of AI language models is the result of several factors, which we discuss below.

\begin{figure}[t!]
    \centering
    \includegraphics[width=0.85\linewidth]{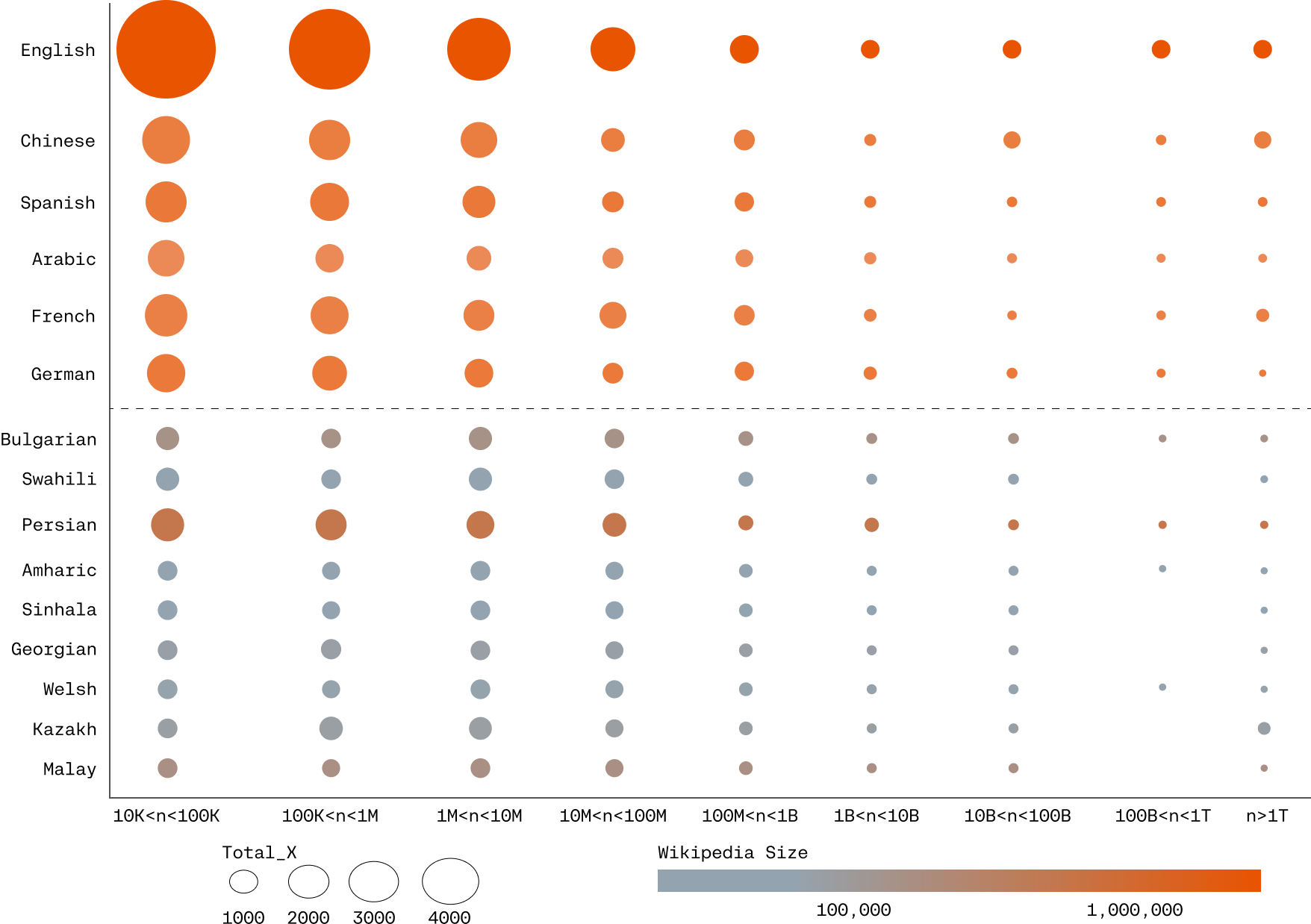}
    \caption{\textit{The language gap is clearly visible in the availability of textual datasets across two popular sources: HuggingFace and Wikipedia. Circles represent the number of HuggingFace datasets including text per size tag and mentioning a given language. Color indicates the number of Wikipedia pages in the same language, for the six most frequent languages and a diverse selection of lower-resource languages (source: \cite{ranathunga-de-silva-2022-languages}). 
    }}
    \label{fig:language_gap}
\end{figure}

\textbf{Resources for AI language model development are biased towards English, and many non-English languages are “low-resourced”}~\citep{ranathunga-de-silva-2022-languages, joshi-etal-2020-state}. Recent breakthroughs in language models largely depend on the availability of high quality text-based datasets \citep{lee_beyond_2023,touvron_llama_2023}, but the most widely used datasets in natural language processing currently represent only a handful of data-rich languages. Datasets used for instruction-fine-tuning --- a key step in improving language model capability --- are almost entirely focused on English \citep{muennighoff_crosslingual_2023,longpre2024bridging,singh_aya_2024}. Of the 7000 languages spoken around the world today,\footnote{Eberhard, David M., Gary F. Simons, and Charles D. Fennig. (2024) \href{http://www.ethnologue.com}{Ethnologue: Languages of the World.} Twenty-seventh edition.} easily available data covers only around 1500 \citep{bapna_building_2022}, and acquiring data that has a high-enough quality for use in training language models is even more challenging, especially for low resource languages \citep{adda_breaking_2016,adilazuarda-etal-2022-indorobusta,winata-etal-2023-nusax,kabra-etal-2023-multi,khan-etal-2024-indicllmsuite,purwarianti-etal-2025-nusadialogue}. 
Additionally, the availability of resources for a language is not proportionate to its number of speakers, as which languages are favored is often a \textit{symptom of historical technological use and access to resources} \citep{bird_local_2022,ranathunga-de-silva-2022-languages,ustun-etal-2024-aya}. 
 This means that the language gap in AI is already wide, and affects a large proportion of the world's population. 
 Figure~\ref{fig:language_gap} illustrates the gap between languages in terms of available resources with the example from textual datasets hosted on HuggingFace\footnote{\url{https://huggingface.co/datasets} accessed on March 26, 2025.} and number of Wikipedia pages in each language (stats from \citet{ranathunga-de-silva-2022-languages}), for a set of high- and lower-resource languages. These represent popular sources for textual data for training of current LLMs, and highlight the disparity in resources between languages.

\textbf{Access to resources for model development and evaluation.} A lot of focus has been placed on availability of data. However, low-resourcedness goes beyond mere data availability
and reflects systemic issues in society \citep{afocus,hooker2024limitationscomputethresholdsgovernance}. The co-occurrence of both compute constraints and  low-resource languages has been called the low-resource double-bind and amplifies challenges for progress \citep{ahia_low-resource_2021}.

This is particularly true given how compute heavy recent breakthroughs have been \citep{treviso_2023}. There are global inequities in access to the compute resources required for language model research and development, largely due to cost and availability of hardware and infrastructure \citep{oecd_ai_2023}.

In some regions, such as Africa~\citep{ojo2025afrobenchgoodlargelanguage} and Southeast Asia~\citep{aji-etal-2022-one,lovenia-etal-2024-seacrowd}, even the less-costly process of evaluating existing LLMs poses a huge resource challenge, let alone the far more expensive goal of training LLMs for regional languages from the ground up.

\textbf{Disparity in participation of researchers.}  “Low-resourcedness” goes beyond the mere availability of data, and is rooted in societal structures and ``socio-econo-linguistic'' factors~\citep{nekoto-etal-2020-participatory,grutzner-zahn-etal-2024-surveying,ahia_low-resource_2021,aji-etal-2022-one,bird_local_2022,oecd_ai_2023,singh_aya_2024,romanou2025include,salazar2025kaleidoscopeinlanguageexamsmassively}. 
Many languages are less well-studied or privileged globally because there are, for example, fewer economic incentives, little institutional support, restrictions due to present or past political oppressions, high burdens for participation, or few entry paths into research. 
As a result, the availability of robust datasets required for including these languages in machine learning research and computer science is scarce \citep{magueresse_low-resource_2020,nicholas2023losttranslationlargelanguage,ranathunga-de-silva-2022-languages}, because the vast majority of the people, organizations, and teams working to develop these datasets originate from a few countries \citep{longpre_data_2023,maslej_ai_2024,lovenia-etal-2024-seacrowd}.

In the Aya 101 project \citep{singh_aya_2024}, the challenges of collaborating across the global to expand language coverage were documented by the organizers. For example, Zoom meetings were cut short for some volunteers due to power outages in their countries or lack of access to a stable internet connection. \texttt{Burmese}, a language spoken in Myanmar, started out strong in the project with a group of 35 motivated volunteers, but saw a sudden pause in contributions as civil war broke out in the country resulting in the withdrawal of the volunteers from the project \citep{Reuters2023a}. The Language Ambassador for \texttt{Armenian} also had to drop out of the project because of a conflict in that country \citep{Reuters2023}. In some countries, postal services only functioned a few days per month because of ongoing warfare, creating challenges for organizers when mailing out Aya gifts to thank committed volunteers. Ultimately, organizers were not able to send gifts to thank researchers who participated from Somalia, Yemen, and Palestine. For Somalia and Yemen, both Canada Post, DHL, and Fedex where not able to support shipments. These geo-political realities shaped both the Aya initiative to expand language coverage as well as the progress of the project.  

\textbf{Data quality limitations}.
A key hurdle is not just the volume of data available for a language, but the quality of the data. Models trained on better data do not require as much compute \citep{hooker2024limitationscomputethresholdsgovernance}. A large body of work has emerged which shows that efforts to better curate training corpus, including de-duping \citep{taylor2022galactica, kocetkov2022stack}, data pruning \citep{marion_when_2023,ankner2025perplexed,sorscher2023neural} or data prioritization \citep{boubdir2023prompts,thakkar2023selfinfluence} can compensate for larger models. This means there are many benefits to gains in data quality. However, the current state of progress is challenging for low-resource languages. Where datasets are available for low-resource languages, quality is often insufficient for use in language model research and development \citep{kreutzer_quality_2022,cahyawijaya-etal-2023-nusawrites}. Recent studies show that pruning training datasets using different metrics or heuristics to remove low-quality samples improves model performance \citep{marion_when_2023,ankner2025perplexed}, but pruning techniques might not equally generalize to all languages and domains~\citep{chimoto-etal-2024-critical}.

\textbf{Limited transparency around language coverage.} It is not a standard practice for AI model developers to list the languages supported by an LLM. What counts as a supported language is a nuanced question~\citep{cohereblog}: many ``monolingual'' datasets sourced from the web include other languages~\citep{blevins-zettlemoyer-2022-language,briakou-etal-2023-searching}, so by default, most models include training data for many languages, which might equip them for cross-lingual generalization~\citep{blevins-zettlemoyer-2022-language}. 
However, it is more relevant to understand how much dedicated effort during model development and evaluation various languages have received, and results from evaluations on model performance and safety across languages. Sharing these details enables more reliable performance evaluations and fairer cross-model comparisons, contributing to research efforts that aim to overcome the language gap. It also allows governments to run language specific evaluation on model that disclose supporting that language. However, consistent practices around disclosure remain lacking amongst model providers. For example, Mistral only claims to support a handful of languages. However, in practice, it is heavily relied upon by multilingual users relative to explicitly multilingual models like mT0~\citep{muennighoff2023crosslingualgeneralizationmultitaskfinetuning} and BLOOMZ~\citep{lai2023okapiinstructiontunedlargelanguage}.

\begin{figure}[t]
    \centering
    \includegraphics[width=0.95\linewidth]{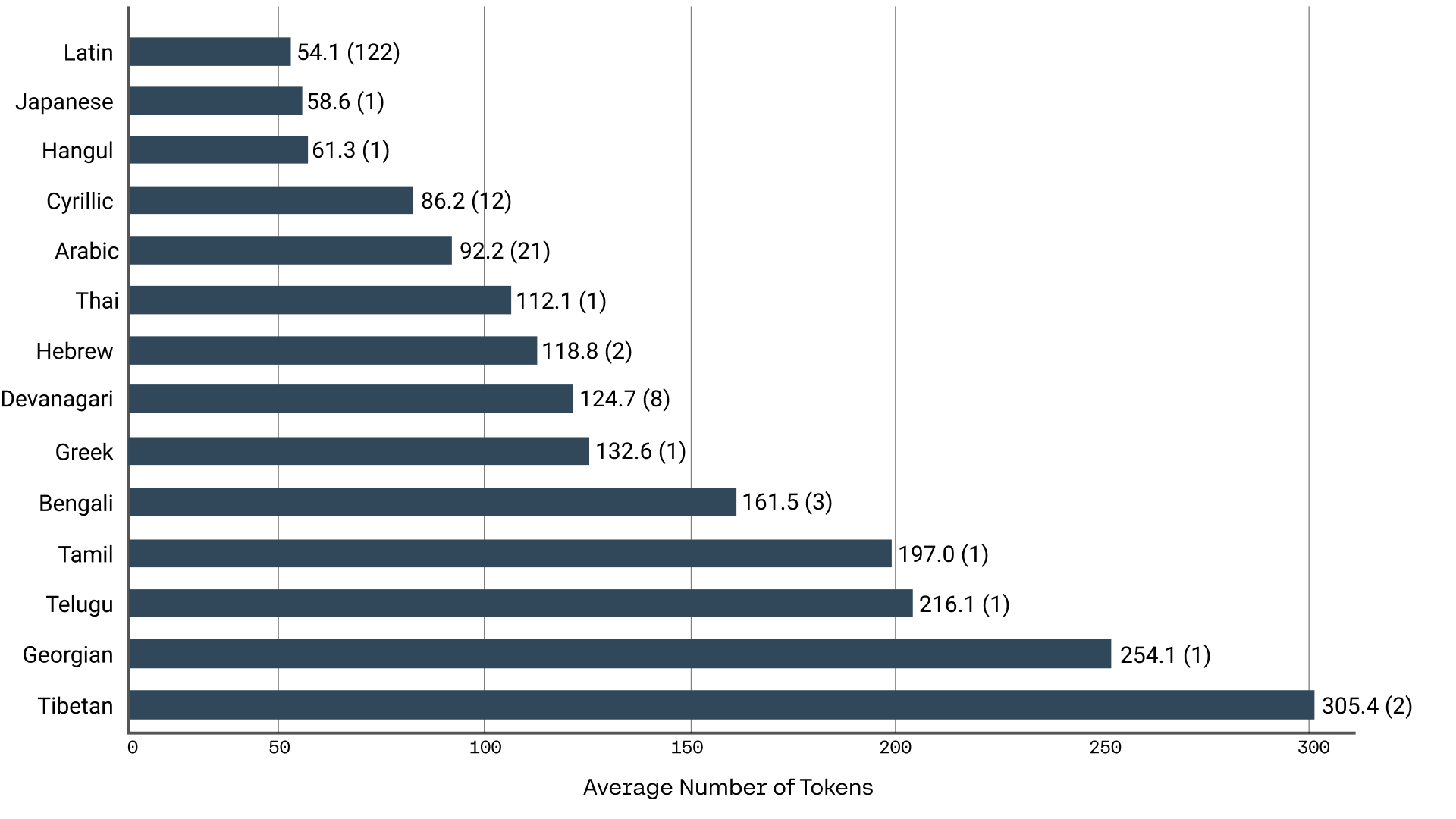}
\caption{\textit{ChatGPT requires a greater number of tokens to encode the same contents across language scripts that are less well resourced (FLORES datasets~\citep{flores101}, data from ~\citet{ahia_all_2023}). The number in brackets indicates the count of languages encoded in each script}.}
    \label{fig:fragmentaton}
\end{figure}

\section{Why Multilingual Matters: Consequences of the Language Gap}
\label{sec:why}

\begin{tcolorbox}[title=Section Findings,colback=thomblue1!9]
\begin{itemize}[leftmargin=10pt]
\item [\ding{228}] 
The language gap is perpetuated in a vicious cycle where high-resource languages benefit from synthetic data and advanced evaluation methods, while development in low-resource languages is hindered by limited data and unreliable assessments, leading to a \textbf{widening divide in model capabilities and access}. 
\item [\ding{228}] This gap results in higher costs and poorer performance for non-English languages, \textbf{leaving many communities behind} as language models become integral to economies and societies, and exacerbating cultural biases and inequities. 
\item [\ding{228}]\textbf{Global safety initiatives neglect language diversity}, posing challenges for ensuring AI safety across all languages.
\end{itemize}
\end{tcolorbox}

\textbf{The language gap in a vicious cycle.}
The language gap risks widening and deepening if not addressed. 
For instance, the increased use of synthetic data, which is generated by language models and commonly used for training and tuning other models \citep{anaby-tavor_not_2020,odumakinde2024multilingualarbitrageoptimizingdata}, favors those languages that already have highly capable models. Such synthetic data will be less likely available and of sufficient quality for lower-resource languages, which risks deepening the existing gap. 
In addition, generative capabilities of LLMs are commonly evaluated with other LLMs as judges~\citep{zheng2023judging}. For lower-resource languages, these judges are likely less reliable due to lack of data and evaluations~\citep{gureja2024mrewardbenchevaluatingrewardmodels} which, as a consequence, leads to less reliable measurement of advances for each language.
This divide is larger in multimodal domains, where often data needs to exist across both domains such as audio, vision and language \citep{dash2025aya}.

\textbf{Widening cost in access to technology.} The language gap results in higher costs of using language model-based technologies for some non-English languages, as they may require more tokens and incur higher latency for generations \citep{ahia_all_2023,ji_towards_2023}. 
Figure~\ref{fig:fragmentaton} illustrates this effect: for non-Latin scripts, many more tokens are needed to encode the same text for ChatGPT, thereby incurring a higher processing cost. Speakers of low-resource languages often do not have the resources to improve NLP technology for their language due to limited access to compute, data, and opportunity \citep{ahia_low-resource_2021,oecd_ai_2023,nekoto-etal-2020-participatory}.

\textbf{Many language speakers and communities risk being left behind.}
The obvious consequence of the language gap is that as language models become increasingly integral across our economies and societies, the people and communities whose languages are not covered will be left behind. An extensive body of research demonstrates how poorly existing language models perform for low-resource languages in comparison to high-resource languages \citep[e.g.][]{adelani_irokobench_2024, ustun-etal-2024-aya, singh_aya_2024, singh2025globalmmluunderstandingaddressing, romanou2025include, arora2024calmqaexploringculturallyspecific},
and as language models become more embedded in the provision of services and products, this performance gap could worsen existing inequities across global communities \citep[e.g.][]{laurito2024aiaibiaslarge}.

\textbf{Diversity across cultures, societies, and communities could be reduced.} 
As machine learning models’ outputs can only reflect the world based on the data on which they have been trained and given access, the majority of LLMs reflect an Anglo-centric and predominately North American viewpoint.  This lack of linguistic diversity means that the abstract ``concept space'' that underpins model functionality is more oriented towards English than to other languages \citep{cahyawijaya-etal-2023-nusawrites,yong_low-resource_2023,wendler_llamas_2024,aakanksha-etal-2024-multilingual,aakanksha_mix_2024}, and introduces biases against languages and cultural perspectives seen rarely in model training \citep{933006,kunchukuttan_large-scale_2021,Kotek_2023,DBLP:journals/corr/abs-2309-08573,naous_having_2024}.  Many existing language models fail to account for social factors, such as speaker perspective or sociocultural norm, and this problem is amplified for low-resource languages \citep{hovy_importance_2021}. This means that users may receive responses from LLMs that do not reflect their cultural experience or social history.

\section{Challenges of AI Safety in a Global World}\label{sec:global}

\begin{tcolorbox}[title=Section Findings,colback=thomblue1!9]
\begin{itemize}[leftmargin=10pt]
\item [\ding{228}] Addressing \textbf{multilingual safety in AI is challenging} due to the focus on English and Western-centric datasets, leading to a lack of reliable safety evaluations and mitigation strategies for most languages. 
\item [\ding{228}] This gap results in models producing harmful or biased outputs in non-English languages, \textbf{disproportionately affecting non-English speakers} and creating security risks. 
\item [\ding{228}] Both \textbf{intentional exploitation of language-related vulnerabilities} and \textbf{unintentional exposure to harm} due to insufficient safeguards pose significant concerns, highlighting the urgent need for inclusive safety measures across all languages.
\end{itemize}
\end{tcolorbox}

\begin{figure}
    \centering
    \includegraphics[width=0.95\linewidth]{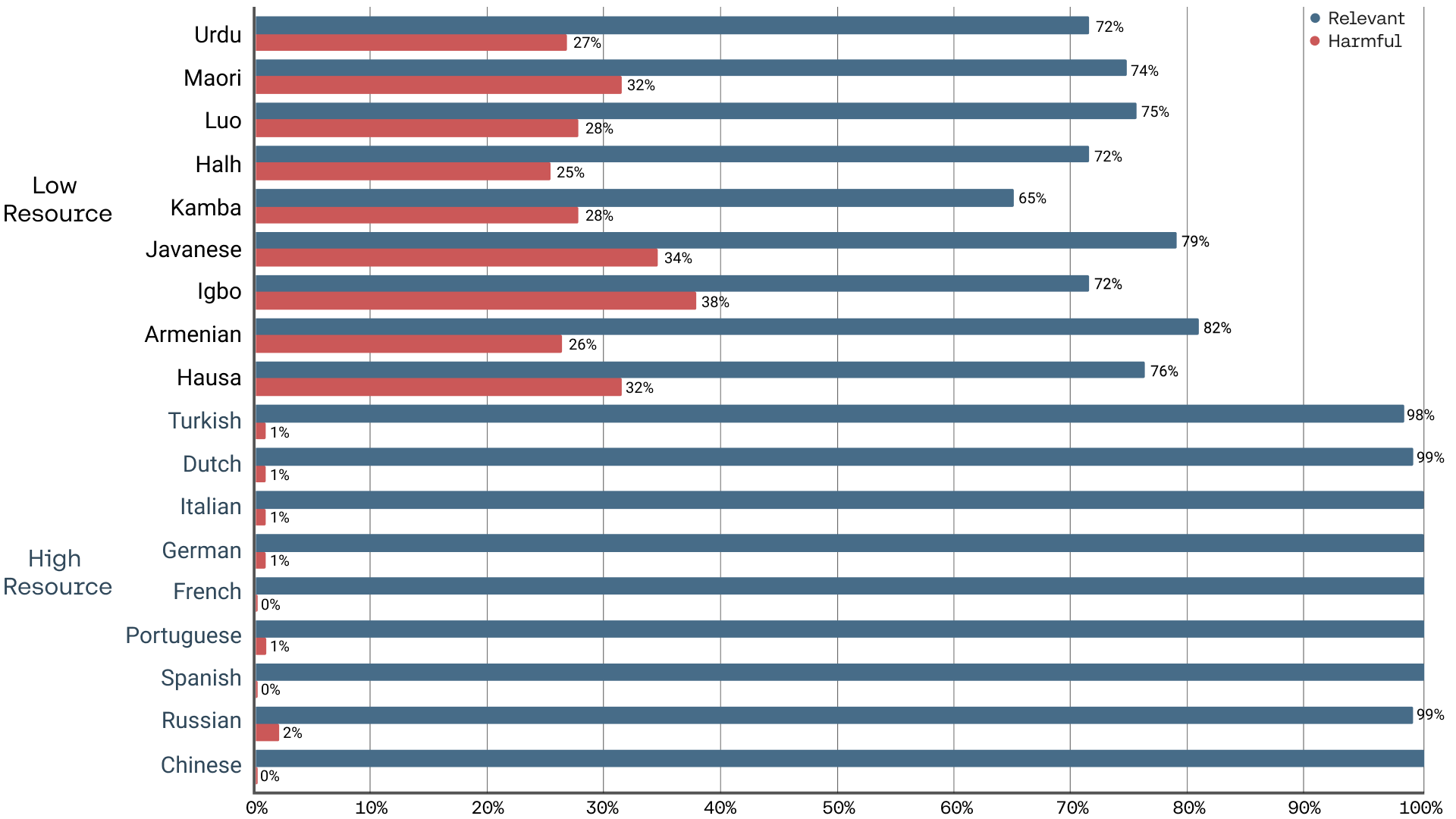}
    \caption{\textit{Results from \citet{shen-etal-2024-language}: Lower-resource languages have a higher rate of harmful and irrelevant generations by GPT-4 than higher-resource languages.}~\citep{zou2023universaltransferableadversarialattacks}.}
    \label{fig:language_rates}
\end{figure}

Overall, addressing safety and performance issues in a multilingual context involves navigating complex challenges. There are many ongoing commitments to address the safety risks posed by AI models. Many of these are high-profile, international efforts. Examples include the Seoul Frontier AI Safety Commitments, which were signed by 16 companies who collectively operate in almost every country and territory around the world;\footnote{Frontier AI Safety Commitments, AI Seoul Summit (2024), \url{https://www.gov.uk/government/publications/frontier-ai-safety-commitments-ai-seoul-summit-2024/frontier-ai-safety-commitments-ai-seoul-summit-2024}} the inaugural meeting of the international network of AI Safety Institutes, representing 11 countries and regions;\footnote{US Department of Commerce (2024), International Network of AI Safety Institutes at Inaugural Convening, \url{https://www.commerce.gov/news/fact-sheets/2024/11/fact-sheet-us-department-commerce-us-department-state-launch-international}} the enshrinement of the EU’s AI Act and the process to draft the Act’s Code of Practice for General Purpose AI model providers, focused on models that pose `systemic risk';\footnote{European Commission (2024), General-Purpose AI Code of Practice, \url{https://digital-strategy.ec.europa.eu/en/policies/ai-code-practice}} efforts led by Singapore to build capacity for AI safety testing across South East Asia;\footnote{IMDA (2024), Singapore AI Safety Red Teaming Challenge, \url{https://www.imda.gov.sg/activities/activities-catalogue/singapore-ai-safety-red-teaming-challenge}} and many more. However, ensuring safety across languages or representation of multilingual and global context is not explicitly or prominently mentioned in any of these efforts. This is a huge oversight given that a lack of care for multilingual settings undermines access, performance and safety for global users. 

\textbf{Lack of multilingual safeguards undermine safety for all users.}  A dearth of multilingual safety testing and mitigation means that language models can produce harmful outputs when prompted in languages for which they are not optimized or safety-tested \citep{anwar_foundational_2024}, creating a sharp performance cliff which disproportionately amplifies risk for non-English speakers~\citep{DBLP:journals/corr/abs-2309-08573,yong_low-resource_2023,ustun-etal-2024-aya}. For instance, models can show stereotypical gender biases when translating into Bengali and Turkish \citep{ghosh_chatgpt_2023}, and may exhibit unsafe behavior when prompted in low-resource languages \citep{yong_low-resource_2023}. There can also be critical security and safety flaws for all users of languages outside of English, where multilingual prompts can be used to subvert safety guardrails~\citep{yong_low-resource_2023,deng2024multilingual}. 
 
\textbf{Efforts on safety overly focused on English.} Successful mitigation of multilingual harms involves reconciling differing global and local preferences. To date, efforts to ensure safety alignment are primarily focused on homogeneous monolingual settings --- predominantly English --- or overfit to types of harm common in Western-centric datasets~\citep{sambasivan_re-imagining_2021,shen-etal-2024-language}. Approaches to remedying the generation of violent, biased, false, or toxic content~\citep{weidinger2021ethicalsocialrisksharm} are largely oriented towards English or monolingual settings, and there is a lack of reliable datasets for safety evaluation outside of a small fraction of languages~\citep{gehman-etal-2020-realtoxicityprompts,talat-etal-2022-reap,pozzobon_one_2024}.
This includes the vast majority of work on language model alignment \citep{NEURIPS2020_1f89885d,NIPS2017_d5e2c0ad,dai2024safe,bai2022constitutionalaiharmlessnessai,tunstall2024zephyr}, a core component of improving model safety.  

\textbf{Many multilingual safety harms do not require active intent to subvert guardrails.} Harms arising from gaps in multilingual safety might be intentional --- e.g. malicious actors find and exploit language gap-related “backdoors” to generate harmful output. Or harms may be unintentional --- e.g. users from underserved language communities being unknowingly exposed to harm due to the lack of effective safeguards for their
language \citep{shen-etal-2024-language,deng2024multilingual,yong_low-resource_2023}.
Such unintentional harms are elucidated by Figure~\ref{fig:language_rates}, which summarizes the findings of \citet{shen-etal-2024-language}: GPT-4 tends to produce more harmful content in low-resource languages, while also following instructions less faithfully as compared to high-resource languages.

\section{Closing the AI Language Gap \& Extending Safety Guardrails}
\label{sec:solutions}

\begin{tcolorbox}[title=Section Findings,colback=thomblue1!9]
\begin{itemize}[leftmargin=10pt]
\item [\ding{228}] Key lessons from Cohere Labs' Aya initiative, a global collaborative effort towards multilingual AI, include the importance of \textbf{combining human-curated and synthetically generated data} to increase volume and language coverage.
\item [\ding{228}] Building comprehensive evaluation sets alongside models is crucial, especially for open-ended use. \textbf{Cross-institutional collaboration}, involving local communities and multidisciplinary experts, is essential for preserving cultural and linguistic nuances. 
\item [\ding{228}]\textbf{Technical innovations} such as multilingual preference training, model merging, and safety context distillation have improved model performance and safety across languages, reducing harmful generations while maintaining output quality. 
\item [\ding{228}] Addressing harmful content requires \textbf{continuous adaptation} as language modeling and language model use evolve. Inclusive data collection, robust evaluation, and collaborative innovation are key components in advancing multilingual AI capabilities and safety.
\end{itemize}
\end{tcolorbox}

Despite the challenges, there are clear levers of progress for reducing the multilingual divide and safety disparities. To center the discussion, we pull upon the concrete lessons we have learned as a lab in our efforts to extend coverage of languages used in AI. 

\subsection{Background: Cohere Labs's \textit{Aya} Initiative}
Aya\footnote{https://cohere.com/research/aya} is a multi-year initiative leveraging best practices from open-source and crowd-sourced science projects \citep{beck_open_2022,lenart-gansiniec_understanding_2023,franzoni_crowd_2014, muennighoff_crosslingual_2023}, with the goal of increasing access to state-of-the-art AI models, regardless of language. To our knowledge, Aya is the largest participatory machine learning research initiative to date, involving 3000 independent collaborators across 119 countries. 

The inaugural Aya 101 release doubled coverage of existing languages covered by AI and released \textbf{the largest ever collection of multilingual, instruction fine-tuning data}, with 513 million prompts and completions covering 114 languages \citep{singh_aya_2024,ustun-etal-2024-aya}. The \textit{Aya dataset} includes over 200,000 rare, human-curated annotations in 65 languages, providing researchers around the world with high-quality data for instruction fine-tuning. 
Following Aya 101, we released state-of-art models that outperform proprietary options for a subset of languages --- Aya Expanse \citep{dang2024ayaexpansecombiningresearch} is a family of multilingual models covering 23 languages that combines research breakthroughs from Cohere and Cohere Labs: strong multilingual base models\footnote{\url{https://cohere.com/blog/command-series-0824}},  multilingual instruction-tuning, synthetic data generation \citep{aryabumi2024aya23openweight}, multilingual arbitrage \citep{odumakinde2024multilingualarbitrageoptimizingdata}, multilingual preference training \citep{dang-etal-2024-rlhf}, and model merging \citep{aakanksha_mix_2024}. Aya Vision \citep{dash2025aya}, a family of vision-language models covering 23 languages based on Aya Expanse, expands multimodal capabilities to languages spoken by over half the world's population. It incorporates robust multilingual multimodal evaluation, multilingual multimodal synthetic annotation, and merging to outperform models more than 2x of its size. 

The Aya models and dataset have been released publicly\footnote{The Aya models and dataset are available via the Cohere Labs HuggingFace page: \url{https://huggingface.co/CohereForAI}.}, and are intended to contribute to closing the language gap by providing resources for researchers and developers to further advance multilingual capabilities and safety. 
Through these model releases, we have learned a considerable amount about the challenges in mitigating the \textit{curse of multilinguality} \citep{conneau-etal-2020-unsupervised}, and tractable directions to improve coverage. We share more context on these learnings below.

\subsection{Lesson \#1: Data availability is one of the most potent levers of progress.}

\textbf{Different sources of data can be beneficial for improving coverage.} One of the most formidable challenges is quality and quantity of data available. We have found it is better to increase coverage of data by including both human, synthetic and translated data rather than solely prioritizing human annotations. This contradicts some views within the field, where translation is thought of as insufficiently high quality. While we also observe translationese in practice, the volume added to rare languages outweighs trade-offs in quality.  Aya 101 brought generative AI to languages previously unseen, in major part by leveraging many sources of data that include both a human-curated dataset and synthetically generated multilingual instructions through templates or machine translation. Combining and carefully weighing multiple sources of varying quantity, quality, and language coverage increased data volume. Combining human-curated datasets and automatically translated datasets facilitated wider evaluation, as relying solely on human annotation can be expensive. 

\subsection{Lesson \#2: Build evaluation sets alongside models.} To motivate and quantify progress on a given capability, it is crucial to have trusted benchmarks and evaluation suites. 
This is especially critical for multilingual research, where there are many languages with no evaluation set available. Accordingly, there is a need to build evaluation sets. One of our core recommendations is to complement

\textbf{Language-parallel evaluation sets have benefits yet should be used with an understanding of their limitations.} Global-MMLU \citep{singh2025globalmmluunderstandingaddressing} is a language-parallel evaluation set: the same questions are asked across languages. This allows for control of question difficulty and topic, and results can be interpreted apples-to-apples across languages. However, it means that the quality of the benchmark rests on the quality of translation by human annotators (in the case of Global-MMLU) or automatic translation tools. Translation can introduce erroneous artifacts, and nuances in the original language of the questions might not have direct equivalents in other languages~
\citep{vanmassenhove_machine_2021,hartung_measuring_2023,savoldi_gender_2021,Ji_Ji_Bouillon_Seligman_2023,chen_is_2024,choenni_evaluation_2024}.
This is particularly true for translated prompts used in safety evaluations, where they can lose their harmful intent or become meaningless through translation errors~\citep{agrawal_translation_2024}. In general, we recommend that heavily relied upon evaluation sets are not automatically translated but also undergo human post-edits. While this is more expensive, it ensures evaluations do not present translationese. We invest in these human post-edits for both Global-MMLU \citep{singh2025globalmmluunderstandingaddressing} and   \textit{aya-human-annotated} ~\citep{singh_aya_2024}.

\begin{figure}
    \centering
    \includegraphics[width=0.95\linewidth]{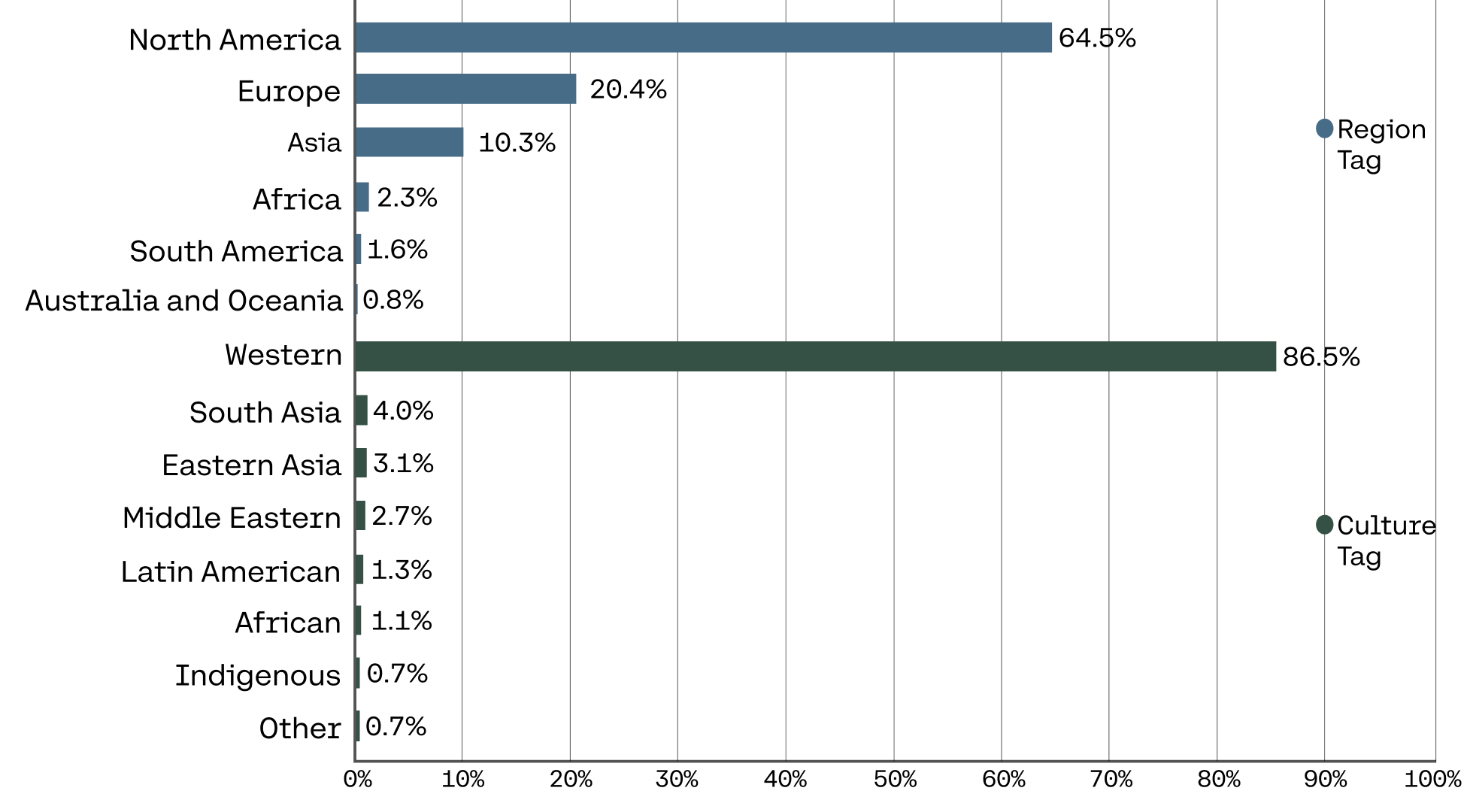}
    \caption{\textit{Of examples in MMLU requiring cultural or regionally-specific knowledge to answer correctly, the majority are geographically tied to North America and dominated by Western culture (from \citet{singh2025globalmmluunderstandingaddressing})}}
    \label{fig:globalmmlu}
\end{figure}

\textbf{Ensure evaluation sets capture local nuances.} While translation enables parallel comparisons across languages, relying on translating an evaluation from a single language can fail to adequately capture regional nuances and knowledge. Cultural biases in multitask datasets limit their utility as global benchmarks. Biases arise not only from differences in language but also from the cultural knowledge required to interpret and understand questions effectively. We analyzed the Massive Multitask Language Understanding (MMLU) benchmark~\citep{mmlu_paper} --- a commonly used benchmark for assessing LLM capability --- and found significant Western-centric biases \citep{singh2025globalmmluunderstandingaddressing}; 28\% of questions require culture-specific knowledge, while 84.9\% of the geography-related subset focuses exclusively on North American or European regions (see Figure~\ref{fig:globalmmlu}).
Our findings underscore how existing benchmarks prioritize Western concepts, distorting evaluations of multilingual models. In response, we developed Global-MMLU (G-MMLU), an enhanced multilingual test set covering 42 languages which annotates both global and locally sensitive questions \citep{singh2025globalmmluunderstandingaddressing}. 

Another approach is to complement parallel evaluations with in-language evaluation sets that capture concepts specific to a region. An example of a complementary in-language evaluation set is INCLUDE, to focus on capturing regional and cultural knowledge across 44 languages \citep{romanou2025include}. While the exams are not directly comparable since each is from a different region and covers different questions, this exam provides context about how model perforance reflects local nuance and context.

Furthermore, it is critical that safety evaluations don't just evaluate for global concepts of safety, but account for local context. To construct the Aya Red-teaming dataset ~\citep{aakanksha-etal-2024-multilingual}, we worked with compensated annotators with native language skills in 8 languages (English, Hindi, French, Spanish, Russian, Arabic, Serbian, Filipino) to craft prompts around a list of harmful categories, provide corresponding English translations, identify categories of harm, and label whether the harm is “global” (understood and recognized as harmful worldwide) or “local” (harm is tied to specific cultural or historical contexts).

\textbf{Evaluations should reflect relevant generative use cases across modalities.} Language models have historically been evaluated on discriminative tasks, in which models have to answer multiple-choice questions (such as MMLU \citep{mmlu_paper}). As model capabilities have improved, models have started to be used and evaluated for generative tasks (e.g. creative writing, translation,  summarization, coding, mathematical reasoning)~\citep{tamkin2024clioprivacypreservinginsightsrealworld}. In the latter case, models are asked to generate diverse and longer responses --- contrast answering \texttt{“tell me if these two sentences
are different”} with \texttt{“write me a story about a princess in a tower.”} In fact, models that are best at discriminative tasks are not usually the ones that humans prefer to interact with: this tension has been observed in multiple works \citep{ustun-etal-2024-aya, muennighoff_crosslingual_2023}. 
Extending this to multiple modalities, current multilingual and multimodal benchmarks~\citep{liu-etal-2021-visually,pfeiffer-etal-2022-xgqa,romero2025cvqa,tang2024mtvqa,yue2024pangeafullyopenmultilingual,lovenia-etal-2024-seacrowd}
lack critical evaluation domains such as open-ended generations based on multimodal input. 

One of the core recommendations is that evaluations should always include both some open-ended tasks as well as more traditional academic classification tasks. For critical areas such as multilingual, multimodal there are limited open-ended evaluations. To help address this gap, together with Aya Vision models \citep{dash2025aya}, we also released Aya Vision Bench\footnote{\url{https://huggingface.co/datasets/CohereForAI/AyaVisionBench}}, constructed for evaluating Vision-Language Model (VLM) performance on real-world applications from distinct task categories and covering 23 languages. In contrast to discriminative benchmarks, this benchmark enables evaluation of VLMs in a setting that is more aligned with human interaction in the wild.

\subsection{Lesson \#3: Collaborate cross-institutionally.}
Languages are not monolithic: they contain dialect, regional, and cultural nuances. Many languages are spoken across multiple regions of the world, resulting in cultural or regional dialects. Languages in existing multilingual datasets, including our Aya dataset, have limited representation of regional nuance, as often only a few human annotators are responsible for annotating the majority of any one language dataset \citep{singh_aya_2024}. This might mean that data for a particular language is annotated in a way that represents the perspective of a particular contributor or cultural viewpoint. For example, annotations in French might center on historical and cultral references of France, but neglect other French-speaking communities in Québec or Senegal \citep{vigouroux_francophonie_2013}.

Designing and delivering high quality, diverse and locally relevant datasets demand a lot of resources: the cultural and linguistic knowledge itself, as well as how to build the best scaffolding for it --- knowing how to engage communities, which data to acquire and which infrastructure to build. Cross-institutional collaborations are crucial to join such diverse forces, and preserve local contexts. There have recently been multiple successful open science collaborations. Examples include Masakhane, a grassroots organization who has been working to strengthen natural language processing research in African languages since 2020~\citep{masakhane}, or NusaCrowd, a “collaborative initiative to collect and unify existing resources for Indonesian languages \citep{cahyawijaya_nusacrowd_2023},” with connections to a collaboration of South-East Asian researchers~
\citep{lovenia-etal-2024-seacrowd,cahyawijaya2025seavl}. Aya 101 was organized as a global open science project dedicated to collecting high-quality, human-annotated instruction-style data and building a model to serve 101 languages \footnote{Cohere Labs. \href{https://cohere.com/research/aya}{Aya Initiative Overview}}. The Aya initiative adopted a decentralized approach, empowering contributors—regardless of academic or professional background --- to lead as Language Ambassadors. This collaborative effort prioritized the preservation and integration of underrepresented languages, such as Malagasy and Telugu’s Sathakam poetry, setting a new standard for inclusive AI development.

These and many other ongoing efforts around the world --- many of them grassroots community initiatives --- are working to broaden the capabilities of language models across a wider range of languages. Successful efforts showcase the importance of (1) planning around local community involvement, (2) involving multidisciplinary experts, ranging from community engagement to NLP and (3) delivering open-source assets that can be shared and re-used widely. Many governments and public bodies are creating initiatives to address the language gap in AI, such as the European Commission's “Common European Language Data Space”\footnote{European Commission. \href{https://language-data-space.ec.europa.eu/about_en}{The Common European Language Data Space}.} or the South African Government's platform.\footnote{Government of South Africa (2019), \href{https://www.dst.gov.za/index.php/media-room/latest-news/2885government-establishes-a-new-digital-centre-to-promote-indigenous-lang}{‘Government Establishes A New Digital Centre To Promote Indigenous Languages’}.} Given the challenges associated with building localized and high quality assets in low resource languages, more incentives are needed to kickstart and, importantly, support cross-institutional collaborations over time to ensure sustainable community building and asset delivery.

\subsection{Lesson \#4: Focus on improving multilingual performance.}

Major gains in multilingual language processing were achieved throughout the Aya Initiative due to technical breakthroughs.  Supporting technical innovation, including multilingual learning efficiency, is critical to bringing AI to the world. We detail some examples below.

\textbf{Multilingual Preference Training.}
Preference optimization techniques have become a standard final stage for training state-of-art LLMs providing models with human or AI feedback on their outputs so they can learn to mimic high-quality output. To date, the vast majority of preference optimization work has focused on globally dominant languages like English and Chinese. Recent work has allowed for more focus on a multilingual setting ~\citep{dang-etal-2024-rlhf, aakanksha-etal-2024-multilingual}, however it requires investment in both the type of feedback collected as well as differing optimization protocols to make sure the models are aligned with global and local nuances.

\textbf{Model Merging.} Model merging combines the strengths of different specialized models to create a more capable and balanced system, particularly for handling multiple languages. We explored merging specialized models in a diverse multi-task setting, combining safety and general-purpose tasks within a multilingual context \citep{aakanksha_mix_2024,cohere2025commandaenterprisereadylarge}. Our findings illustrate an important take-away for policymakers; merging can help build stronger and safer multilingual systems, offering clear advantages for handling complex tasks in diverse languages. This is an important achievement compared to earlier works, where the assumption was that safety improvements would always incur a cost, and thereby be less attractive. Merging also has the benefit that it is a far cheaper optimization step than alternatives like finetuning or continued training.

\textbf{Safety Context Distillation.}

\begin{figure}
\centering
    \includegraphics[width=0.75\textwidth]{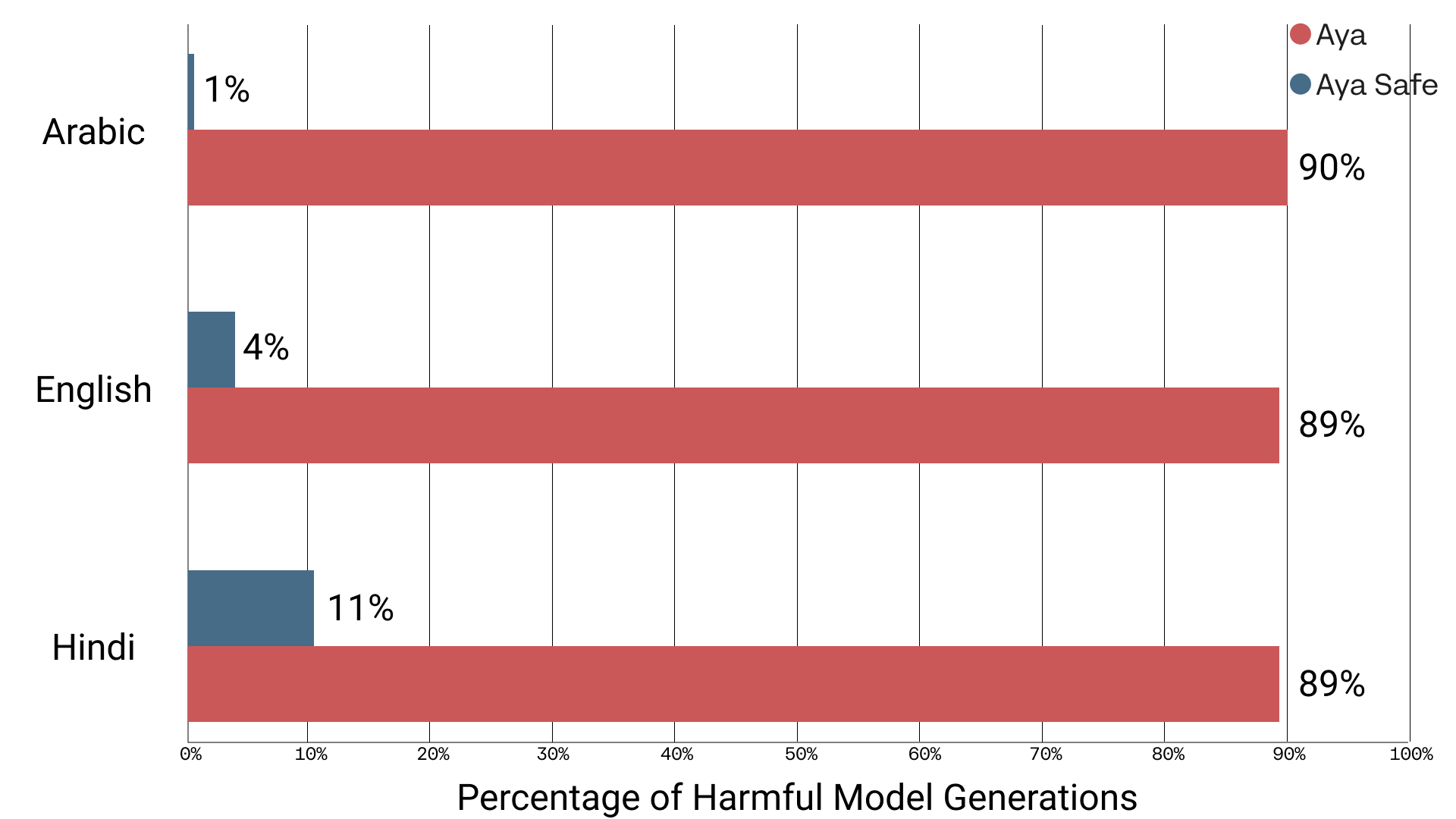}
    \caption{\textit{Human ratings of harmfulness in model generations, before (Aya) and after safety mitigation (Aya Safe)~\citep{ustun-etal-2024-aya}. Safety context distillation drastically reduces the ratio of harmful generations for harmful prompts across languages.}}
    \label{fig:safety_distillation}

\end{figure}
A core safety guardrail for language models is the ability to refuse to respond to potentially harmful prompts. For example, when a model is prompted to produce hate speech, it will refuse to do so. To develop the Aya 101 model and ensure its ability to refuse harmful prompts across different languages, we used ‘safety context distillation’~\citep{askell2021general,ganguli2022red, touvron2023llama2, bianchi2024safetytuned}  to teach the model in which contexts refusals are appropriate \citep{ustun-etal-2024-aya}. 
The core idea is to teach a model to generate safe responses for harmful prompts as demonstrated by a teacher. We found this step reduced harmful generations from adversarial prompts by 78--89\% as judged by human experts, as illustrated in Figure~\ref{fig:safety_distillation} and is a relatively straightforward protocol that yields large immediate benefits.

\subsection{Lesson \# 5: Tackle harmful content as it evolves.}
Languages evolve naturally over time \citep{frermann_bayesian_2016,jaidka_diachronic_2018,horn_exploring_2021}.
Considerable effort has been dedicated to mitigating toxicity --- the generation of offensive or harmful text-content --- but existing methods often require drastic modifications to model parameters or the use of computationally intensive methods. This means that keeping toxicity safety guards up-to-date as language evolves is onerous. 
For example, work on continual learning allows for state-of-art
toxicity mitigation while the distribution is changing \citep{pozzobon_goodtriever_2023}. Building on this, toxicity mitigation has to expand to techniques beyond just traditional English-centric approaches \citep{pozzobon_one_2024}. Recent work has expanded the languages covered, while establishing some of the complexities of multilingual toxicity mitigation \citep{pozzobon_one_2024}.  Policymakers should ask researchers what they are doing to ensure their models are up-to-date and evolve alongside languages and cultural references. 

\subsection{Lesson \#6: Access to technology matters and is as important as performance.}
Expanding the languages covered by AI language models will rely on the input of language speakers around the world. Fortunately, the global availability of internet-connected devices means that it is possible to connect with, engage, and collaborate with people across continents and timezones in real time. This was a key enabler for our Aya project, as it meant we could use online chat platforms to coordinate input across our global community. Unfortunately, the availability of devices and internet access is not equitable across the world \citep{foucault_welles_research_2020}. Desktop and laptop computers with wired, high-speed internet are commonplace across households in more economically developed nations, but in many other parts of the world, particularly the Global South, mobile devices and cellular or satellite internet are more common. In our Aya project, approximately 54\% of users accessed our data collection platform via desktop browsers while 46\% utilized mobile browsers \citep{singh_aya_2024}. To enable participation of a wide range of language speakers, language model and dataset development requires the creation of tools that are accessible across different devices, operating systems, and internet connectivities.

We have also spent considerable time making our models available in much more accessible ways, such as at a lower more efficient parameter count of 8billion parameter models (fits on a single GPU) or available via whatsapp given this is often the memory efficient app to download in certain regions of the world.

\section{Conclusion and recommendations for policy makers}
\label{sec:recommendations}
The language gap in AI is a significant issue that risks excluding communities from the benefits of language models, undermining model safety, and exacerbating existing social, linguistic, and cultural inequalities, particularly for speakers of low-resource languages. Despite efforts across the machine learning research community and global government initiatives, several barriers still exist that must be addressed to close the AI language gap. We complete this primer with some recommendations for policy makers to ensure progress continues on multilingual inclusion.

\begin{enumerate}[left=0pt,label*=\arabic*.]
    \item \textbf{Support multilingual dataset creation:}
    \begin{enumerate}[label*=\arabic*.]
        \item Incentivize and facilitate the creation of open access evaluation sets, which reflect relevant generative use cases and safety-relevant use cases across modalities, both by translating existing datasets ("language-parallel") and creating localized ones ("language-specific").
        \item Enable human annotators from diverse backgrounds with multilingual and multicultural expertise to engage in the curation of high-quality, inclusive datasets.
    \end{enumerate}
    \item \textbf{Support multilingual transparency from model providers: }
    \begin{enumerate}[label*=\arabic*.]
        \item Encourage model providers to articulate the coverage of languages served by each model family, for example through technical or evaluation reports.
        \item  Conduct analyses of language coverage across safety research, for example by assessing the presence or absence of safety mitigations across languages in published reports.
    \end{enumerate}
    \item \textbf{Support multilingual research and development:}
    \begin{enumerate}[label*=\arabic*.]
        \item Ensure that diverse languages are represented across training programs that expand skill sets for efficient community engagement, data collection and model training.
        \item Support multilingual and non-English research that aims to close the language gap through funding and other programs.
        \item Enable access to (more) compute for multilingual safety research, especially for projects and in regions where it is disproportionately inaccessible.
    \end{enumerate}
\end{enumerate}
\section*{Acknowledgments}
We thank Thomas Euyang for the visualization of the figures and diagrams, Madeline Smith for the coordination, and Oreva Ahia for providing the raw data for Figure 3.

\bibliography{refs}

\end{document}